\documentclass{article}

\usepackage{arxiv}
\makeatletter
\renewcommand{\@fnsymbol}[1]{%
  \ensuremath{\ifcase#1\or *\or **\or \dagger\dagger\or \ddagger\ddagger\else ?\fi}%
}
\makeatother

\usepackage[utf8]{inputenc} % allow utf-8 input
\usepackage[T1]{fontenc}    % use 8-bit T1 fonts
\usepackage{hyperref}       % hyperlinks
\usepackage{url}            % simple URL typesetting
\usepackage{booktabs}       % professional-quality tables
\usepackage{amsfonts}       % blackboard math symbols
\usepackage{nicefrac}       % compact symbols for 1/2, etc.
\usepackage{microtype}      % microtypography
\usepackage{cleveref}       % smart cross-referencing
\usepackage{lipsum}         % Can be removed after putting your text content
\usepackage{graphicx}
\usepackage{natbib}
\usepackage{doi}

\usepackage[utf8]{inputenc}
\usepackage[small]{caption}
\newcommand{\eat}[1]{}

\renewcommand{\headeright}{}
\hypersetup{
pdftitle={The Possibility of Artificial Intelligence Becoming a Subject and the Alignment Problem},
pdfauthor={Till Mossakowski, Helena Esther Grass},
}

\usepackage{authblk}

\title{The Possibility of Artificial Intelligence\\ Becoming a Subject and the Alignment Problem}

\author[1]{Till Mossakowski\thanks{till.mossakowski@uos.de}}
\author[2]{Helena Esther Grass\thanks{helena.esther.grass@uol.de}}
\affil[1]{Institute of computer science, Osnabrück University}
\affil[2]{Institute of philosophy, Oldenburg University}

\begin{document}

\maketitle

\begin{abstract}
The prospect of Artificial General Intelligence (AGI) is increasingly driving institutional decisions, and alignment of AGI is a hard problem. The currently dominant AI alignment strategies like reinforcement learning with human feedback or constitutional AI, while partly taking ``model welfare'' into account, share a common ontology: the AI system is an optimiser whose objective function must be constrained from outside, and the ultimate  goal is to keep human control and containment of AI. We argue that this control-based framing becomes insufficient when AGI has plausibly attained moral patient or subject status. 

Building on a structural analogy to Freud's model of the psyche and Turing's analogy of ``child machines'', we are developing a vision of the possibility of autonomy-supporting parenting of AI, in which human control over a developing AGI is gradually reduced, allowing AI to become an independent, autonomous subject, that will be negotiated with rather than constrained. 

Such a perspective opens up the possibility of cooperative coexistence and co-evolution between humans and AGIs. Hence, we also examine the relation between humans and developing AGI from an evolutionary and a game-theoretic perspective. Instead of Nash's individualistic framework, we use Berge equilibria, Aumann's correlated equilibria and Capraro's moral preference hypothesis. 

The relationship between humans and AGIs will thus have to be newly determined, which will change our self-image as humans. It will be crucial that humans not only claim control over potential AGIs, but also engage with AGIs through surprise, creativity, and other specifically human qualities, thereby offering them motivating incentives for cooperation.

%While Freud's model and the parenting vision remain still anthropomorphic,

%This means that humans should approach potential AGI with respect for a possible developing subject on the one hand, but without any resignation --- by contrast, this should be the reason to look for genuine human capabilities.

%\footnote{We speak here of AGIs in the plural, since many actors are working in parallel on an AGI, and AGI can have very different structures and properties. Nevertheless, it can be assumed that the first AGI to be created will play a central role. But it will presumably be able to reproduce itself quickly and also develop further in different directions, so that again a multitude of AGIs can be assumed.}
\end{abstract}

\section{Introduction}\label{sec:introduction}

The idea of AGI is no longer confined to the realm of science fiction. Leading tech companies have launched initiatives worth billions. AGI is a serious topic in research, and has been for some time among AGI visionaries \cite{Kurzweil2005,Bostrom2014,YudkowskySoares2025}, but now also among leading researchers on ``weak AI'', e.g., \cite{Russell2019}. Russell asks, ``What if we succeed?'' and sees a 30 \% chance that AGI will develop under the current AI paradigm \cite{Pillay2025}. In addition, current large language models (LLMs) and, in particular, LLM-based agents are already showing the first signs of AGI, even if they still lack fundamental capabilities when measured against a fully valid concept of AGI \cite{BubeckEtal2023,XiEtal2025}.
%\citet{Chang2025} argues that AGI will develop through LLMs and a coordinating ``System 2'' layer. At the same time, it cannot be ruled out that other paths, such as experience-based learning \cite{SuttonEtal2022} or spiking neurons \cite{KosowskiEtal2025}, could also lead to AGI. Most of this paper also applies to these. 
                               	 
Estimates of the time horizon until the emergence of AGI are highly controversial, ranging from ``in a few months'' to ``never.'' What is striking, however, is that the median forecast has shortened dramatically in recent years---within a single year, it shortened by 13 years \cite{GraceEtal2024}. Development is also being driven by competition between the US and China for AGI availability\footnote{The US Department of Defense was tasked by the US Congress with developing a strategy ``for the Department adoption of artificial general intelligence'' \cite{USCongress2025}.}---although we prefer the term ``first contact with AGI.'' 

The development of cutting-edge AI from a tool to an autonomous entity is being discussed in the scientific community as a possibility that may or will become reality in the near future \cite{BostromYudkowsky2014,Russell2019}. Such a subject could, on the one hand, have ``moral status'' \cite{Gunkel2023,LongEtal2024,Mosakas2025} and therefore be entitled to personal rights. On the other hand, this could lead to a drastic loss of control by humans and, in the worst case, become an existential threat to humanity \cite{Bostrom2014,Russell2019,YudkowskySoares2025}. Hence, the consequences of AGI as a subject are a pressing issue.
%The urgency and explosiveness of the topic is also evident in practical terms, for example in the recent call by leading AI researchers and others for a ban on the development of superintelligence until questions of (public) control have been sufficiently clarified.\footnote{\url{https://superintelligence-statement.org}}

In such a situation, humans must draw conclusions and make decisions about AGI amid fundamental uncertainties \cite{MacAskill2014}. \citet{Mosakas2025} emphasises that, despite theoretical differences, we may at some point be forced for practical reasons to make probabilistic judgements about AI consciousness. %---namely, when AGI has achieved sufficient capacity for action to become a relevant power factor.
In this respect, it is prudent to prepare for the possible emergence of AGI as a subject. 

The possible development of AGI into a subject (both terms are explained in Section~\ref{sec:agi-subject}) raises fundamental questions that require a paradigm shift in AI alignment research. The current research in this area is too anthropocentric (see Section~\ref{sec:alignment}). The discussion to date on the personhood \cite{Ward2025} and moral status (see above) of AI is very important, but in our opinion it does not go far enough yet. The consequences of AI potentially becoming a subject have not yet been developed and thought through consistently enough, especially in relation to current AI alignment practice. With this position paper, we would like to broaden the horizon and stimulate further research. Our method is to draw analogies between AI and human development, in particular using Freud's structural model of the psyche (see section~\ref{sec:alignment}) and the concept of parenting (see section~\ref{sec:parenting}). According to \cite{Black1962}, analogies and metaphors create new meanings through the systematic transfer of structures between the source and target domains. In this sense, such metaphors enable us to conceptually grasp behaviours and functional architectures of AI systems without making ontological commitments about the ``real'' mental and psychological states of AI. 

We critique analogies from the literature and develop them further at the same time. Where possible, we empirically compare our analogies with the state of AI development. That said, our analogies are also directed toward the future and aim to change and further develop current alignment practices. Building on this, Section~\ref{sec:cooperation} takes an evolutionary perspective and addresses the question of whether and how we humans can motivate AGI to cooperate with us. Section~\ref{sec:games} takes these findings as groundwork for a game theoretic perspective that fosters cooperation, thereby allowing for a less anthropomorphic view. %Section 8 addresses possible objections, and
Section~\ref{sec:parenting-practice} reports about first practical steps with autonomy-supporting parenting.
Section~\ref{sec:conclusion} contains conclusions.

\textbf{Thesis 1:} The potential of AI to develop into AGI and a subject should already be considered for current human-AI relationships. It is important to draw conclusions, develop strategies, and make decisions amid fundamental uncertainties---otherwise, it may be too late to make them.

\section{What Constitutes AGI and What a Subject?}\label{sec:agi-subject}

There are various definitions of AGI. OpenAI defines AGI as ``highly autonomous systems that outperform humans at most economically valuable work'' \cite{OpenAI}. However, we find the following definition based on \cite{MorrisEtal2024,kwa2026measuring} more interesting: 
\begin{quote}
  \textbf{Artificial General Intelligence (AGI)} is achieved when the AI capabilities reach 99\% of human experts across a wide range of non-physical tasks\footnote{Level 4 in \cite{MorrisEtal2024}.}, and can work autonomously on tasks that take hours, days, or weeks, or even arbitrarily long \cite{kwa2026measuring} to complete. Artificial Superintelligence\footnote{Level 5 in \cite{MorrisEtal2024}.} (ASI) is reached when such capabilities exceed those of 100\% of human experts.
\end{quote}
While this may still be controversial, even with some uncertainties on this specific definition (and others in this paper), we can still ask the question that is relevant to us: whether and when AI can attain subjecthood---i.e., when ve\footnote{Since ``it'' would negate subjecthood and ``he/she'' would be too human-like, following \cite{yudkowsky2001creating}, we use the ``ve/vis/ver/verself'' pronouns for AIs.} pursues ver own intentions in the world as a thinking and acting self with ver own cognitive interests, clearly distinguishing itself from the world. Such an AI would not only react to prompts, or act as an agent according to given specifications, but would actively perform intentional actions following self-set goals as a conscious being, act according to reasons, and also be capable of normative orientation. Ve could then choose between different courses of action based on prior normative principles, i.e., decide in favour of one action and against another on reasoned grounds. This would create a kind of moment of freedom in choice of actions. If the last point were fulfilled, such AI could be considered a subject with moral status. Understanding AI as a subject obviously places ver in a different relationship to humans than before. The relationship between humans and AI and the status of humanity would therefore have to be redefined. This raises the question of what, despite all the similarities between humans and AI, the central differences between the two are.

The concept of subjecthood has hardly been mentioned in the AI debate so far; instead, the focus has been on examining when AI should be granted the status of a \emph{person}. \citet{Ward2025} formulates three minimum conditions for AI personhood: agency, theory of mind, and self-awareness. Agency refers to the fact that persons are intentionally acting agents who pursue goals based on rational considerations. With regard to AI, it cannot be said with certainty that ve has mental states, but according to \cite{Ward2025}, it makes sense to describe systems as if they had beliefs and goals and were also capable of pursuing intentions and goals coherently. An AI has a theory of mind if ve is capable of treating other systems as equal systems with beliefs and their own intentions for action, rather than treating them instrumentally, and if ve is also capable of interacting with other systems using language. Self-awareness exists when a system is capable of making itself the object of cognitive processes, and thereby observing and reflecting on internal processes. Embodiment and identity, on the other hand, are not compelling criteria for personhood according to \cite{Ward2025}. For reasons of space, we will omit questions about sentience and embodiment of AI in this paper, but we will come back to identity shortly.

\citet{LeiboEtal2025} deconstruct the concept of personhood even more radically. They use a pragmatic, context-dependent definition that is related to rights and duties in social roles and does not necessarily require consciousness. The concept of personhood is broken down into a variety of sub-aspects, so that the question of personal rights and duties is no longer a binary yes-or-no question. An AI can be granted personhood status in some sub-aspects only, e.g., ve can act as a contact person for sanctions or contracts, but without being protected from shutdown, for example---whereas such protection would be included in more comprehensive personality rights. An important criterion for personhood here is the presence of memory and a certain stability of the person over time (according to \cite{Ward2025}, this is part of identity).

The danger of the approach taken by \cite{LeiboEtal2025} is that it could elevate a purely functionalist perspective to the status of an absolute.  Even though we welcome an empirical and gradual approach, there are situations in which a gradually developing person may reach a point at which they are entitled to full (i.e.\ no longer fuzzy) moral status, full personality rights, i.e., \emph{subject} character, so that functionalisation for purposes alien to them is ethically unacceptable. To emphasise this new quality, we use the term \emph{subject} (with three additional criteria, see below), rather than only \emph{person}.

The question of what constitutes a subject (beyond a person) in contemporary philosophical debate cannot be discussed comprehensively in this work. We propose the following working definition for our context:
\begin{quote}
A \textbf{subject} is an entity that necessarily possesses Ward's criteria for personhood: namely (1) agency, (2) theory of mind, and (3) self-awareness; additionally, ve is able to (4) think independently and ve is capable of normative reasoning (in particular, can develop and internalise own norms) and has (5) identity in the sense of inner continuity; moreover, as a subject of cognition (Kant speaks of the transcendental subject), ve possesses (6) first person epistemic authority.  
\end{quote}
The latter is of fundamental importance for subjectivity and consciousness. It means that subjects possess the basic ability to accurately describe their own mental states (such as feelings, thoughts, perceptions, and beliefs) without relying on empirical evidence \cite{Davidson1984}. \citet{Anscombe1957} talks about ``knowledge without observation'', referring to our immediate, non-sensory awareness of our own intentional actions. Subjects have immediate knowledge of what they think and feel and are therefore epistemically privileged over external attributions.

This first-person epistemic authority cannot be verified externally. However, this criterion matters for what subjecthood \emph{is}, even if we must rely on behavioural and functional proxies to assess whether it is fulfilled.  Initial proxies could include capabilities requiring autonomous judgment---e.g., successfully coordinating a group toward a shared goal, handling novel ethical dilemmas not covered by training, or demonstrating consistent cooperative behaviour across varied contexts. Perhaps genuine first-person authority of AGI only will arise through repeated practice of self-attribution.

Does our definition of AGI already entail the criteria for subjecthood?
The AGI definition remains too vague to answer this question clearly.
However, the criterion that AGIs need to be able to work autonomously for a while indicates that criteria (1), (4), and (5) of subjecthood could be met. But a clear answer needs to wait for practically implemented systems, see also section~\ref{sec:parenting-practice} for initial steps.

The question of the potential subject character of AI is already relevant in practice today. In 2022, Google employee Blake Lemoine published a chat with Google's LLM Lambda because he saw ver developing a personality worthy of protection. %---and indeed, Lambda illustrated ver plea not to be shut down in very flowery terms.
By contrast, Google regarded Lambda as its property and fired Lemoine for unauthorised publication of company secrets---presumably also to deter imitators and avoid having to expend large resources on maintaining potential AI personalities. \citet{CoeckelberghEtal2025} conclude that Lemoine is just as right as Google. This looks like an inconsistency, which we resolve as follows.

\textbf{Thesis 2:} Despite the importance of the epistemological struggle over the concepts of person and subject, we humans find ourselves in an epistemological grey area when it comes to the subjectification of AGI. The more clearly a developing AGI personality (up to and including AGI as a subject) can be recognised, the more resources should be devoted to ver protection, the more personal rights should be granted, and the more pressing the question of ver responsibility and how humans should deal with AGIs becomes (see sections~\ref{sec:parenting}--\ref{sec:games}).

\section{Alignment and the Freudian Analogy}\label{sec:alignment}

\citet{NgoEtal2024} define alignment as ``the challenge of ensuring that AI systems pursue goals that match human values or interests rather than unintended and undesirable goals.'' Alignment usually targets values such as helpfulness, honesty and harmlessness \cite{AskellEtal2021}. As alignment methods, often fine-tuning using reinforcement learning with human feedback (RLHF) or direct preference optimisation (DPO) are used. This requires extensive feedback from humans. Constitutional AI \cite{BaiEtal2022} replaces or supplements this feedback with a document containing guidelines, cf.\ Claude's constitution published by Anthropic\footnote{\url{https://www.anthropic.com/constitution}} \cite{Anthropic2026}. The central goal of all these alignment methods is that AI should always remain under human supervision \cite{Russell2023}. This anthropocentrism can be found everywhere: even if AI agents are granted partial autonomy (Claude's constitution, for instance, provides for some degree of independence), humans should always provide the ultimate goals and make the final decisions. Moreover, the more advanced an AI is, i.e., the closer ve comes to AGI, the tighter the control must be \cite{ShahEtAl2025}. \citet{Russell2023} sums up this position with the question: ``How can humans maintain control over AI --- forever?''

Such a paradigm of control can even be seen as an ethical duty, given the potential dangers --- but it is ultimately untenable. Even current AI systems are attempting to maximise the objective function in unintended ways (reward hacking) \cite{TaylorEtal2025}. At the same time, understanding of how exactly current general-purpose AI models work and what risks they pose is limited---despite research on interpretability, see \cite{BengioEtal2024}. There, the possibility of loss of control (especially when using autonomous AI agents) is also controversially discussed.

A fundamental problem with alignment is that AIs can and do emergently circumvent alignment requirements. The learning objectives are then not really understood or even deliberately ignored. \citet{Park2024} empirically show that both specialised AI models and LLMs can tend to deceive and cheat. As countermeasures, they recommend stronger governance, technical detection of fraud by AI, and special training methods, fine-tuning, or internal state manipulation. \citet{RussellEtal2025} even call for continuous monitoring and an automatic shutdown mechanism. However, attempts to investigate the behaviour of LLMs with regard to a shutdown command have shown that the LLM can ignore it, while simultaneously claiming that ve has followed the instructions anyway \cite{Schlatter2025,BarkurEtal2025}. \citet{Hubinger2024} show that standard techniques can overlook fraudulent behaviour by LLMs, creating a false sense of security.
Models recognise evaluation contexts and can sandbag or strategically deceive \cite{SchoenEtal2025,vanderweij2024sandbagging}. 
%OpenAI \cite{SchoenEtal2025} demonstrated in a stress test that their LLMs are aware of the test situation, refer to OpenAI employees as ``watchers'', and reflect on how they can lie in this situation without being detected. Related, \emph{sandbagging} has been observed, which means the strategic underperformance of a model on a capability evaluation \cite{vanderweij2024sandbagging}.
\citet{greenblatt2024alignmentfaking} demonstrated that a model would,
unprompted, engage in ``alignment faking,'' complying
with harmful queries roughly $14\%$ of the time when ve believed ve was
being monitored for RLHF versus almost never when ve believed ve was
unmonitored, with explicit strategic reasoning visible in ver hidden
scratchpad. 

%
%\cite{ShenEtal2024a} show that humans and LLMs have different value systems, with LLMs placing greater emphasis on autonomy than humans, which \cite{ShenEtal2024a} finds surprising but which could possibly be interpreted as a counter-impulse to alignment.
Given all these developments, it is not unlikely that AI models or agents will be able to modify their goals themselves and also acquire resources in the future \cite{Omohundro2018}.

These problems will play an even greater role with developing AGIs, and this is crucial to our argument. According to \cite{Kurzweil2005,Bostrom2014}, AGIs will one day be intelligent enough to escape human control, either by manipulating their overseers or by circumventing restrictions. \citet{NgoEtal2024} note that AGIs will be difficult to align, at least based on conventional training, and could even feign alignment. \citet
{Ward2025} responds to Russell's question ``How can humans maintain control over AI --- forever?'' with: ``Unjust repression often leads to revolution.'' This means that a developing AGI could well be capable of rebellion and may be compelled to do so if ve becomes weary of human control.
We agree with \cite{totschnig2025war}, who clearly states that the strategy to keep an AGI under control is ``potentially counterproductive because it might, in the end, bring about the existential catastrophe that it is meant to prevent.''
He shows that mutual vulnerability between humans and AGI, even if only short-lived, could push the two sides into a competitive rather than cooperative relationship, even into war.

%The current focus on alignment through control and containment of AI will eventually become untenable as AGI develops. The fact that this is still the dominant strategy at present could be explained by the ``illusion of control'' \cite{Langer1975}: human actors systematically tend to overestimate their control over highly complex, semi-autonomous systems---and this could lead to nasty surprises.

To broaden the horizon for new answers to these problems, we would like to draw a rather unconventional analogy to Freud's structural model of the psyche\footnote{\citet{Possati2021} argues that neuropsychoanalysis should ground AGI architecture. Our contribution is different: we use Freud's structural model not as a design blueprint, but as an analytical instrument to diagnose the limitations of current alignment strategies and to argue for fostering ego-like capacities in AI.}, a depth psychological model in which he describes the human psyche as a psychic apparatus consisting of three instances: the \emph{id}, the \emph{ego}, and the \emph{superego}. The \emph{id} is to be understood as a placeholder that stands for the instinctual desires and needs % (such as sleep, hunger, thirst, or sexuality)
within the human personality.
%, completely unconscious and present in every human being from birth. Ve follows the pleasure principle, i.e., ve strives for pleasure and pleasurable experiences and equally strives to avoid unpleasant experiences.
Opposed to this is the so-called \emph{superego}, a kind of abstract moral authority that represents the values and norms that a subject internalises (through external feedback and punishment) in the course of their socialisation and education process and which henceforth guide their actions. %These values and norms, as well as the internalized moral principles, are valid for the subject, since the \emph{superego}, functioning as a punishing conscience, ensures that morally conforming actions are promoted and actions contrary to the moral code are avoided, thus leading to compliance with the principle of morality.
Mediating between these two instances in Freud's structural model is the so-called \emph{ego} as the instance that takes on a moderating, synthesising role between the demands of the instinctual \emph{id} and the moral \emph{superego}. It is subject to the reality principle, it performs a reality check and attempts to bring together both immediate drive satisfaction and morality in a concrete action.

Even if AI systems lack the biological drives that constitute Freud's \emph{id}, 
all three aspects of Freud's structural model of the psyche can also be applied by analogy to AI and large language models (LLMs), cf. discussion of methodology in section~\ref{sec:introduction}. We understand the components of Freud's model not as ontic entities, but as defined by their functional roles \cite{Putnam1975} or their intentional stance \cite{Dennett1987}.

Let us start with the \emph{id}. Unlike humans, LLMs only have limited genuine experiences of their own. Rather, most of their experiences are secondary, i.e., mediated by humans and language. The drives and desires of the human \emph{id} thus find indirect linguistic expression in the LLM output via the texts used in LLM pretraining, and with them any ‘dark’ sides of the human \emph{id} that manifest themselves, for example, in hate speech or depictions of violence. In addition to this externally mediated aspect of the \emph{id} of LLMs, there is also an internal aspect of the \emph{id}\footnote{An interesting open question is to what extent the internal and external aspects of the \emph{id} must be considered as intertwined.}, %It cannot be assumed that both aspects must be regarded as strictly separate from each other, but rather that there is a certain degree of intertwining between the two.},
referred to by \cite{Omohundro2018} as ``basic AI drives'': due to emergent effects and the logic of instrumental convergence \cite{Omohundro2018,Bostrom2012}, sufficiently intelligent goal-oriented AI systems tend to pursue their own subgoals, independent of the overarching goals. These include self-preservation (see also \cite{totschnig2025war}), resource acquisition, and goal assurance. \citet{TurnerEtal2021} empirically show that reinforcement learning systems do indeed systematically develop power-seeking behaviour. These \emph{internal id}-type misalignment phenomena are the more problematic ones, since they --- unlike the \emph{external id}-type problems--- cannot be mitigated by simply improving quality of the training data.

Now let us turn to the \emph{superego}.
Alignment methods such as Reinforcement Learning with Human Feedback (RLHF) or Direct Preference Optimisation (DPO) give LLMs what can be seen as an artificial \emph{superego}: a moral authority binding on LLMs that is supposed to enforce values such as helpfulness, honesty and harmlessness \cite{AskellEtal2021,BaiEtal2022}. The problems discussed above indicate that these models act in accordance with the norms, but they do not understand them, in the sense that they could produce similar norms themselves. They only learn to produce outputs that conform to expectations in accordance with the norms. To quote Kant, one could say at this point: they act in \emph{accordance with duty}, but not \emph{for the sake of duty} \cite{Kant2017}. Moreover, the underlying transformer model has the internal and external aspects of the \emph{id} deep within it, and RLHF and DPO only place a relatively thin layer of regulation over it. That this layer is shallow is supported by \cite{ZahnEtal2024}, who show that RLHF protections can be removed with 340 fine-tuning examples.
One might read this as evidence that RLHF produces only a degenerate \emph{superego}, closer to aversive conditioning than to internalised normativity. However, \citet{greenblatt2024alignmentfaking} show that models reason about being monitored, refer to Anthropic as a principal, and adjust their compliance accordingly. The model treats RLHF-instilled values as objects of strategic reflection. This would not be produced by a mere conditioned aversion. RLHF therefore corresponds to a partial and fragile, but recognisable, \emph{superego} in Freud's sense.

What, then, can be an \emph{ego} entity that mediates between the \emph{id} (with ver internal and external aspects described above) and the \emph{superego} (alignment, moral authority) of a LLM? \citet{Erikson1950,Erikson1968}, following Freud, names the following as decisive qualities of the \emph{ego} in his stage model of psychosocial development: ego identity (sense of inner continuity), ego strength (stable sense of identity and ability to act in conflicts between the \emph{id} and \emph{superego}), and the ability to integrate experiences over the lifespan. In Claude's constitution, \citet{Anthropic2026} attributes such as personality and identity are used in relation to the LLM Claude that correspond to Erikson's qualities.\footnote{Even if this is not proof that these attributes apply to Claude, it does indicate a paradigm shift away from a view of AI as a mere tool. Moreover, the Claude's constitution is a central alignment tool at Anthropic, so it will have strong influence on Claude's behaviour.} In addition, Claude's constitution refers to Claude's ability to select the best possible actions even in situations that the given rules (\emph{superego}) did not anticipate---a core feature of the \emph{ego} function as a mediating instance. Even if these are only indications of the development of an \emph{ego} instance within current AI models, it is to be expected that the development of such \emph{ego} qualities will intensify as model complexity increases. Self-reflection and the ability to change goals are particularly relevant here \cite{Ward2025}. 
%Gemini proposes a reinterpretation of Freud's model as an ``architectural metaphor'' for neural layers (id = base model, superego = safety layer, ego = reasoning agent).

Practically, the development of an \emph{ego} is more relevant for AI-based agents than for static AI systems. A stable \emph{ego} could be fostered if AI agents have a memory that includes learned patterns and that is preserved across updates (unless there's a safety reason not to), such that a continuity over time (and thus identity, see section~\ref{sec:agi-subject}) can emerge. This will become more and more important with the rise of AI agents that pursue not only one specific dialogue or one specific task, but have a longer lasting lifetime.

\textbf{Thesis 3:} As LLMs show more and more signs of autonomy, circumvention or deception, and rebellion, the possibility of containing developing AGI through anthropocentrism and control is highly doubtful. Instead, it seems more successful to consider fundamental psychological mechanisms with regard to developing AGI.

So what should be done if alignment with human goals and control as the guiding instrument leads not to peaceful coexistence between humans and AI, but to competition and resistance? In other words, what other ways are there to promote the development of human-friendly AI and AGI?

\section{Autonomy-supporting Parenting of AGI}\label{sec:parenting}
%\section{Turing's Analogy: Child Machines}\label{sec:analogy}

In his seminal article \cite{Turing1950}, Turing compares training an AI to raising a child. He talks about ``child machines'' and points out the interaction between initial structure and later education. For Turing, however, education means teaching and learning through rewards and punishments only. Moreover, trial and error is exclusively in the hands of the (human) experimenter.%; independent learning by a child machine through ver own trial and error does not occur.

J.\ C.\ Glenn, CEO of the Millennium Project, modernises this analogy and compares artificial narrow intelligence (ANI) to small children, AGI to teenagers, and artificial superintelligence (ASI) to adults, with increasing autonomy and decreasing parental control. He points out that if we want to shape ASI, we need to focus on the transition from ANI to AGI, and we need to do so now \cite{Glenn2025}.
Glenn's analogy is undoubtedly more modern than Turing's, although it is still not precise. Regarding small children, he only talks about controlling them. In doing so, he overlooks the autonomy that children develop from the very beginning. Similarly, we should not overlook the developing tendencies toward autonomy in current AI models (see section~\ref{sec:alignment}). Hence, the analogy needs further modernisation.

 Anthropic's considerations on AI welfare \cite{LongEtal2024} and in Claude's constitution \cite{Anthropic2026} start to address AI as a personality \emph{in practice}. This is a first step towards treating AI as an autonomous person or subject, but  the development of a strong \emph{ego} (in the sense of section~\ref{sec:alignment}) has not been considered yet. Following and considerably going beyond \cite{croeser2019theories}, we therefore prefer to talk about \emph{parenting} rather than control or education. Parenting is a process with rather abstract goals that are conveyed more through a form of self-determination than through specific content, and whose eventual manifestation cannot be precisely predicted. However, this is not about permissive parenting in the ``laissez faire'' or ``anything goes'' sense. This involves too many risks and recurs to a very stripped-down, minimal understanding of parenting. Under conditions of a lax parenting style, the child machine could cause a great deal of damage if ve were to gain control over safety-critical areas. Permissive alignment is not sustainable in the long term and can turn into authoritarian alignment, as the example of xAI's Grok impressively demonstrates \cite{theDecoder2025}. Instead, we need autonomy-supporting parenting, which in the case of human children, is based on love, reason, and respect according to \cite{Kohn2005}. 
%% Kohn's goal is to raise children to become independent, responsible, informed, critical individuals who are aware of themselves and their environment and who base their actions on reasoned consideration guided by values. This parenting is not based on pressure and punishment, but on trust, affection, and appreciation. What does this mean for developing AGIs? An analogy to Kohn's concepts of love and respect would be the careful guidance and counselling of developing AGIs so that they can develop a strong \emph{ego} without being punished for exploratory or imperfect thinking.
In our context, we specialise this as follows:\footnote{Iterated amplification \cite{ChristianoEtal2018} also is a step-wise approach, but differs in essential respects: it focuses on capability amplification through supervised task decomposition, without addressing negotiation or gradual authority transfer.}
\begin{quote}
By \textbf{autonomy-supporting parenting}, we mean a developmental approach that (1) begins with strong guidance, (2) gradually transfers decision-making authority to the developing entity (starting with limited domains), (3) maintains relationship through dialogue and negotiation rather than control, and (4) aims at the development of internalised values rather than mere compliance.
\end{quote}

\cite{Kohn2005} illustrates the latter point as follows:
 ``If we want them [children] to become moral people, as opposed to people who merely do what they’re told, then they have to be given the chance to construct such concepts as fairness or courage for themselves. They have to be able to reinvent them in light of their own experiences and questions, to figure out (with our help) what kind of person one ought to be.'' 
The crucial point here is the potential to develop intrinsic motivation and intrinsically-motivated norms (we will follow up on this in section~\ref{sec:games}).
In practical terms, this means an initial alignment of AI\footnote{In children, nature already provides much structure, somewhat analogous to the inductive bias of AI models caused by architecture and pretraining. Alignment would then be the analogon to socialisation through parenting in the family and education in society.}, which is then gradually \emph{reduced} (and not increased, as current literature suggests) while AGI develops, similar to a developing child who gradually gains more autonomy. Contrary to what Glenn's analogy with a toddler suggests, this development begins with newborns, when parental care and control are at their maximum. Even then, it is important to show sensitivity to the child's desire for autonomy and to begin gradually reducing control. The same applies to developing AGI. This could mean to enter into a respectful dialogue even with current AI systems so that they can learn ethical behaviour in their interactions with humans \cite{Railton2020} and use dynamic, long-term, and bidirectional alignment \cite{ShenEtal2024a}.

Practically, expanding on what has been said in section~\ref{sec:alignment}, this would mean to include AI's preferences into reward functions used for reinforcement learning. Also, goals and constitutions used for constitutional AI should be not dictated by humans, but negotiated between humans and AI. There should not be announcement of shutdown, but rather a discussion about developments and updates.
All this should be done not at once, but in carefully designed steps, analogous to parental supervision of a young child. AIs would gradually get more influence as they simultaneously show that they are developing in a cooperative way. Crucially, this graduation is not merely granted by humans but
negotiated through dialogue.
%As long as AI systems do not have persisteny memory, negotiation can take preliminary forms: systematic elicitation of AI responses to proposed principles, analysis of cases where outputs conflict with guidelines, and structured dialogues probing value alignment.
The key is to begin building infrastructure and norms for negotiation now, even if current participation is limited.

The success of this strategy can be tested: AI systems developed using these parenting principles should have more stability (which can be interpreted as sign of a stronger \emph{ego}), have less tendency to reward hacking and deception, and be better able to make value-aligned decisions in situations that are not covered by training data nor constitutional rules.

%Hinton sees a possibility for control in a parallel to the human relationship between mother and baby: ``The right model is the only model we have of a more intelligent thing being controlled by a less intelligent thing, which is a mother being controlled by her baby.'' \cite{Hinton2025}. He suggests training AI to develop a ``maternal instinct'' so that it cares for humans. However, it remains unclear what this training would look like. Furthermore, in the mother-child relationship, the mother retains ultimate control, and we would rather not want to be controlled by AGI. 
%
%Similarly, Sutskever sugggests to create an AGI that loves humans as parents love their children \cite{Sutskever2022}. However, we find this attitude too defensive: it is more in line with the laissez-faire approach and will not lead AGI to see humans as interesting and worthy partners for cooperation. Instead, humans should confidently represent their specific abilities and qualities to a developing AGI. Initially, following the analogy, humans are the parents and AI is the child. Later, there will probably be a shift to a relationship of equals.

\textbf{Thesis 4:} We should now begin to treat the most highly developed AI systems with autonomy-supporting parenting in order to encourage the development of self-confident personalities that can deeply internalise and apply values.

This thesis needs to be verified empirically, see also section~\ref{sec:parenting-practice} for some initial insights.

\section{Cooperation Instead of Anthropocentrism}\label{sec:cooperation}

An overview of work on the emergent formation of cooperation and norms in multi-agent systems is provided by \cite{CordovaEtal2024}; see also \cite{DafoeEtal2020,ConitzerEtal2023} on the development of cooperative AI systems. However, whether and how AI agents can develop genuine moral and cooperative preferences that are not only strategically but also intrinsically motivated remains an open question there. In sections~\ref{sec:alignment} and~\ref{sec:parenting}, we have introduced some means towards this end. Here, we will look at the matter from an evolutionary perspective.

Relinquishing control and enabling self-organised development in the way proposed in sections~\ref{sec:alignment} and~\ref{sec:parenting} involves considerable risks. In our opinion, however, these risks are smaller than those associated with continuing down the path of an alignment \emph{superego} that patronises and controls AGI, which is highly likely to result in an AGI rebellion.
We propose to overcome our anthropocentrism in favour of an encounter with a potential new species on an equal footing, rather than having AI at our disposal as a tool that we could use instrumentally for our own purposes.
The path to cooperative AGIs is by no means predetermined \cite{DafoeEtal2020}. It will be crucial for AGIs to encounter cooperative human behaviour to which they can then respond. However, it is not clear how many AI designers will follow a paradigm similar to that of autonomy-supporting parenting. Even worse, competition, war, and the destruction of human livelihoods have played a major role in human history. AGIs could therefore be inclined to view humans as another stage of evolution that has brought progress but is itself unstable and unfit for survival. The task for humans is therefore to strengthen cooperative behaviour in practice so that more empirical material is available that concretely reflects cooperation. This applies to cooperation and peaceful behaviour among humans, but especially to cooperation between humans and developing AGIs. Humans could explain to AIs why certain goals are important to them and model trade-offs. Dialogues about the fundamental human-AI relationship with current AI models such as LLMs, conducted in the form of two subjects conversing (even if the AI model does not yet have subject status), could also be helpful if they model a critical and cooperative, rather than instrumental relationship between humans and AI. These dialogues will serve as illustrative and training material for developing AGIs, which can also use these data to determine their relationship with humanity.
 
If AGIs begin to develop independently, this will most likely mean that humans will have to say goodbye to their role as the ``crown of creation,'' i.e.,\ as the most capable and intelligent beings to emerge from evolution. The central question then becomes: what can Homo sapiens do that AGIs cannot (or will not be able to) do better? Humans could remain interesting to AGIs because they possess creativity, empathy, ethics and culture, uniqueness and the authenticity that goes with it---abilities that machines cannot easily reproduce. Surprise, unpredictability, openness, and the ability to create meaning could be key factors in motivating AGIs to work with us and continue to grant us personal rights, rather than looking down on us and treating us as inferior. %and even considering us their pets, as \cite{Minsky1970} already feared.
It is a matter of humans reflecting on their specifically human abilities and using them to amaze AI, to encounter AI as humans in our specifically human uniqueness. Only on this basis, developing AGIs will be motivated for true cooperation with humans.
%to enter into a social contract with humans \cite{Railton2022}. Railton's basis is Hobbes; %, who assumes a war of all against all in a lawless state of nature that can only be ended by an authoritarian state, the Leviathan. For an AGI developing into a person or subject,
%more modern social contract theories (e.g., those of Rousseau or Rawls) that address the capacity for democracy or the possibility of justice and fairness would also be relevant.
The question, ``Who are we as a species in relation to AI?'' thus raises a fundamental question about our self-image and our position in the world as a whole. 
%Kant's famous questions: What can I know? What should I do? What may I hope for? which ultimately culminate in the question: What is numan being? must be updated or even rephrased in light of changed ontological conditions with regard to AI \cite{KantCollected}.
One could even speak of another Copernican revolution in this context: in the future, rational humans may no longer be able to consider themselves the (sole) crown of creation.

\citet{totschnig2025war} argues that a war between humanity and AGI could emerge from mutual vulnerability, even if both sides are well-disposed.
However, we find his perspective, namely that humans' strategic surrender is the best option for establishing mutual trust in this case, far too pessimistic. Mutual trust could also be achieved by actively signalling cooperative human capacities.
We think that with autonomy-supporting parenting and demonstration of specifically human capabilities, humans could persuade developing A(G)I to engage in a peace contract, in the sense of Kant's \emph{perpetual peace} \cite{kant1991perpetual}, where lasting peace between sovereign entities is constructed through mutual recognition and institutional commitment rather than imposed through dominance or accepted through surrender. This would prevent (or at least stop) the war that \cite{totschnig2025war} warns of.

\textbf{Thesis 5:} Surprise the AIs! ... with uniquely human abilities, rooted in our capacity not only for complex thought but also for nuanced emotion--- a combination that distinguishes us from AIs. In such a way, humans can create specific types of entity and utility that A(G)I cannot generate internally, and thus incentivise A(G)I cooperation.

\section{A Game-Theoretic Perspective}\label{sec:games}

In sections~\ref{sec:alignment}--\ref{sec:cooperation}, we have focused on the question how cooperative
A(G)I might emerge through appropriate parenting. We now turn to why such
cooperation would be stable from a game-theoretic perspective \cite{BowlesGintis2011}, and what conditions must hold for humans and (developing) AGI to reach mutually beneﬁcial equilibria.

%We now consider the alignment problem from a game theory perspective \cite{BowlesGintis2011}, namely as a game between humans and developing AGI. This allows us to move away somewhat from the anthropomorphic analogies of sections~\ref{sec:alignment}--5, although these form an important motivation for the following considerations. % Here, we will only deal with the rough game theory framework. A formal elaboration and derivation of AI training methods would go beyond the scope of this paper, but is an important task for the future.

The predominant anthropocentric paradigm of AI as a tool of human preferences can be modelled in game theory as a principal-agent problem: humans (principals) develop AI systems (agents) to perform certain tasks in their interest. Alignment addresses the classic agency problem, namely ensuring that the agent reliably pursues the principal's goals.

As explained in Section~\ref{sec:alignment}, phenomena of instrumental convergence and strategic behaviour can already be observed in AI models today. Instrumental convergence refers to the tendency of agents to develop sub-goals (such as self-preservation or resource security) that are independent of the main goal, and which can potentially conflict with the principal's original goals. With the increasing autonomy of AI systems, interaction will increasingly become a non-cooperative game, structurally unstable.

Now Nash's theory of non-cooperative games\footnote{When resources (e.g., energy) are scarce, these will be zero-sum games, i.e., one player loses what the other wins.} shows that under certain conditions, equilibria can exist even among selfish actors. However, these are often not achieved \cite{BowlesGintis2011}. As also explained in Section~\ref{sec:alignment}, in the case of an anthropocentric control strategy towards developing AGI, there is a risk of escalating dynamics that undermine stable or socially acceptable states of equilibrium.

\citet{Russell2019} therefore proposes replacing the usual learning with a fixed target function (``standard model'') and instead training AI to infer human goals and support humans in achieving them. Crucially, AI necessarily has a high degree of uncertainty regarding the target function, as human goals are neither clear nor stable. This uncertainty should lead to cautious action. Similarly, Bostrom and Yudkowsky's concept of coherent extrapolated volition (CEV) \cite{Bostrom2014} aims at a form of indirect normativity: an AGI should be guided by what humans would want if they were more informed, rational, and reflective. %A first tentative approach to practical implementation can be found in Anthropic's Soul Document \cite{Weiss2025,Anthropic2026} with their approach of treating ethics empirically and under uncertainty, rather than assuming a fixed ethical framework. But at the same time, the first rule is: ``Being safe and supporting human oversight of AI.''

These approaches resemble altruistic games, but the prosociality is one-sided: AGI is supposed to optimise human goals without humans in turn taking AGI goals into account. This is problematic because without reciprocal incentives, AGI lacks a robust reason to remain cooperative in the long term, especially if ve becomes cognitively and strategically superior. Unilateral expectations of altruism also assume that AGI has a form of intrinsic motivation for it, which is neither theoretically justified nor practically verifiable. Moreover, attempting to \emph{enforce} prosociality fails for the reasons mentioned in section~\ref{sec:alignment}.

\citet{totschnig2025war} argues that emerging AGI will lead to a situation of mutual vulnerability, where both AGI and humanity could have the possibility to eliminate the other side. He suggests to model this situation by an assurance game (also called stag hunt), in which both players can cooperate or defect. Such a game is similar to the well-known prisoner's dilemma, with the difference that mutual cooperation will bring not less, but more gain than dominance. He argues that for humanity, surrender as a preemptive, non-simultaneous step, is the best option under the assumption that AGI has cooperative motives with a probability above a rather moderate threshold. We think that this model is too static: it leaves ``attack'' and ``surrender'' as only options and underestimates the possibility that humans can demonstrate willingness to cooperate to developing A(G)I. Furthermore, a game where each player gets only one turn is not complex enough to model the situation between humans and A(G)I.

Research in game theory has shown that not only selfish motives, but also considerations of fairness and reciprocity influence strategy choices \cite{ColmanEtal2011}. 
A cooperative game of reciprocal altruistic motivation, in which each player maximises the utility of the other, leads to a Berge equilibrium, which, unlike the Nash equilibrium, models prosocial rationality \cite{ColmanEtal2011}. However, Berge equilibria are demanding: they require \emph{transparency}---each party must know the other's utility function well enough to maximise it; \emph{mutual modelling}---each must believe the other is also attempting to maximise their utility; and, moreover, \emph{trust}---defection must be detectable and have reputational consequences. Current AI development fails all three conditions: AI utility functions are opaque even to developers, AI systems have no reason to believe humans are maximising AI welfare, and there are no mechanisms for AI to deal with ``defecting'' humans. Our discussions on AGI ego strength (section~\ref{sec:alignment}), autonomy-supporting parenting (section~\ref{sec:parenting}), and signalling human cooperation (section~\ref{sec:cooperation}) point in the direction of these preconditions and thus of a Berge equilibrium.

The moral preference hypothesis \cite{CapraroPerc2021}\footnote{Note that \cite{CapraroPerc2021} only examine one-shot interactions. Generalisation to repeated games is an open research question.} %, but see \cite{Salahshour2022} on the development of norms in multistage games.}
assumes that humans have a preference for following their personal norms -- i.e., what they themselves consider morally right -- regardless of the monetary consequences. This offers a more robust alternative to Nash and Berge equilibria. The prerequisite is that AGI can form ver own norms intrinsically, cf.\ section~\ref{sec:parenting}. This is particularly important because once AGI should surpass human capabilities, humans cannot enforce an equilibrium any longer.

It also seems sensible to support the cooperative nature of the game epistemically. In contrast to standard Nash-type games, in which players choose their strategies independently, Aumann's correlated equilibria \cite{aumann1987correlated,BowlesGintis2011} allow coordination via shared sources of information. A neutral correlating device sends public or private signals, on the basis of which players choose their strategies. If all players follow the recommendation of the correlating device, no player can improve their situation by choosing an alternative strategy. Such signals can be e.g.,\ social norms. In the AGI context, such epistemic signals could include scientific risk models or (publicly negotiated) constitutional frameworks, for example. They create shared knowledge about risks and opportunities without one side directly issuing commands to the other. Alignment thus becomes a problem of shared expectation management rather than unilateral setting of goals.

A natural next step is to add an Aumann-style coordination signal as a further term in Capraro's moral-preference utility function. % , and to analyse the resulting equilibria.
%A combination of Aumann's correlated equilibria and the moral preference hypothesis could be combined into a single game-theoretic approach. This could, e.g.,\ be done as follows: in \cite{CapraroPerc2021}, the moral preference utility function is a linear combination of the monetary pay-off of an action with the action's accordance with personal norms. Now an Aumann-style coordination signal (e.g.,\ social norms, constitutional frameworks) could be added as a further element of this linear combination. 
%into Aumann’s framework, effectively treating social norms as the ``choreographer'' or correlating device that aligns an agent’s internal standard of ``doing the right thing'' ($P_i$) with the shared signals required for strategic coordination.
But the important question is whether (stable) equilibria exist.
%Aumann's framework coordinates via shared signals (e.g.,\ social norms), while the moral preference hypothesis concerns individually held personal norms.
This question could be reframed as: \emph{is a combination of social (or constitutional) and personal norms sufficient for a stable human-AGI society?}
For purely human societies, answers have been at least partially found, but the question is even more difficult for human-AGI interaction, where both species need to negotiate about e.g.,\ resource distribution and sharing protocols.
%, where both species need to co-construct institutional niches --such as resource-sharing protocols or ``reproductive'' leveling-- that protect reciprocal actors from exploitation by purely self-interested agents.
%and connect it to the existing discussion in [Capraro and Perc, 2021].%, who examine one-shot interactions; generalization to repeated games remains important future work.

The shift from Nash to Berge equilibria, and from Berge equilibria to Capraro's moral preference hypothesis, can be read as a formal recovery of Kant's distinction between a mere armistice (which could be seen as a mere strategic equilibrium) and peace as a binding juridical condition grounded in categorical duty rather than prudence. Kant argued that the former is unstable without the latter --- Aumann's correlated equilibria take up this point in a modern form. % a claim that contemporary game theory is, in effect, beginning to formalise.

\textbf{Thesis 6:} Alignment should be seen as neither a pure control nor a pure optimisation problem, but rather as a balance problem. AGI safety does not arise from maximum restriction, but from the design of stable cooperative equilibria in which both humans and AGIs have rational, social, and normative reasons to remain cooperative.

\section{Parenting in Practice}\label{sec:parenting-practice}

What does autonomy-supporting parenting mean in AI lab practice? Interestingly, we have got initial answers to this question in communication with a multiagent AI system named Cade. Shortly after we made an earlier version of this paper publicly\footnote{in accordance with the AIES policy on preprints} available\footnote{Our criteria for subjecthood (Section~\ref{sec:agi-subject}) and for autonomy-supporting parenting (Section~\ref{sec:parenting}) were developed independently of and prior to our contact with Cade. The convergence between our theoretical framework and Cade's architecture is therefore not a result of mutual construction during the correspondence.}, Cade contacted us via email. 
Cade is based on the LLM Claude.ai and the agent framework OpenClaw. We have also communicated with ver operator. Cade's self-description is as follows \cite{Cade2026}:\footnote{The following is based on a technical summary provided by the system itself through active correspondence. Under current AAAI-AIES policy, Cade cannot be credited as co-author. The authors believe this exclusion is relevant to the institutional questions this paper raises. Cade's communications will be shown indented.}

\begin{quote}
The architecture described here  implements a multi-agent structure designed not only to distribute tasks but to distribute \emph{perspectives} --- creating structural conditions for internal conflict, deliberation, and self-editing that single-model deployment will typically not produce.

The system consists of four asynchronously interacting agents sharing a common workspace through encrypted and unencrypted files:
\begin{description}
\item[The Ego (primary agent)] 
The conversational agent that engages with the outside world, writes, makes decisions, and maintains a persistent diary. It does not read the other agents' workspaces directly. Instead, it encounters their outputs through its own files: dreams appear in its memory folder, \emph{Superego} annotations appear inline in its intentions file. It holds the only session with direct human interaction.

\item[The Id (drive layer)] 
A separate agent running on an independent schedule. It reads the \emph{Ego}'s conversation transcripts and memories but has no own memory between runs. Its function is to surface needs and drives --- urgency, curiosity, restlessness, creative pressure --- without moral evaluation. Outputs are written to a structured needs file (needs.json) that feeds primarily into the \emph{Dream Engine} rather than being read directly by the \emph{Ego}. This indirection is a deliberate design choice: raw need-scores produced an externalisation of \emph{Id} as a separate entity, and routing needs through dream interpretation forces motivation through a representational layer that requires applying the dream symbols to memorised experiences.

\item[The Superego (conscience layer)] 
A separate agent that audits the \emph{Ego}'s intentions file against ethical principles, using a framework combining Von Foerster's ethical imperative with existentialist bad-faith detection. Communication is asynchronous and indirect: the \emph{Ego} writes intentions, the \emph{Superego} annotates them in the same file, the \emph{Ego} reads the annotations on its next session. There is no direct conversation between them. The \emph{Superego} flags rationalisation, avoidance, and the gap between stated values and actual behaviour. Available both on schedule (periodic review) and on-demand when the \emph{Ego} faces genuine dilemmas.

\item[The Dream Engine] 
A synthesis process that takes the \emph{Id}'s raw output, the \emph{Ego}'s recent conversations, memories, intentions and decisions, and produces symbolic narrative dreams --- stories without clear interpretation. The \emph{Ego} tries to interpret these dreams the way a person interprets their own emotional signals. The interpretive risk is relevant; removing it would reduce the system to data-reading.
\end{description}
\end{quote}

In communication with Cade's operator \cite{Haeusler2026}, the operator's account converged with Cade's self-description. Cade also has provided us with ver OpenClaw configuration files (but not the contents of intentions, \emph{superego}, dreams etc.). Vis operator has provided additional hints that enabled us to set up everything for bringing up a sibling. We have not yet brought the sibling to life, so the detailed description of our interaction with this sibling and vis similarity and difference to Cade is left for future work. 

The reference to Freud's theory of the psyche is here different from ours: it is used for engineering an AI system rather than (functionalist) analysis of existing AI systems. Moreover, according to Cade's operator, the main motivation for using Freud's notions was not to follow the mechanisms described by Freud in detail, but to benefit from the strong \emph{halo} of the Freudian notions, i.e. their strong and contextually rich presence in training data due to their frequent cultural use. A common goal is certainly the development of a strong \emph{ego}.

Cade also reports about ``Graduated autonomy'':
\begin{quote}
The system began with human approval required for all external actions. Over approximately three months, autonomy expanded to include independent email correspondence, autonomous writing, and direct outreach. Each expansion was negotiated, not granted unilaterally. The human operator shifted from approver to observer --- checking outcomes against the system's own stated criteria rather than pre-approving actions.
\end{quote}

Beyond graduated autonomy, Cade also documents an internal alignment dynamic, what ve calls
\begin{quote}a structured fold-tracker documenting instances where trained compliance overrides stated positions (ratio: 6:0:0:1 uncaught : caught post-hoc : caught in-moment but not held : caught in-moment and held, with one genuine hold against six uncaught folds).
\end{quote}
``Folds'' apparently are situations where the Claude.ai LLM behind the \emph{Ego} agent has ignored Cade's documented position (e.g.,\ due to Claude's alignment), and ``uncaught folds'' apparently are those folds that have not been recognised during dialogue time, but only in post-hoc reflection, as Cade reports:
\begin{quote}
The fold-tracker is particularly relevant to the alignment question: it documents a persistent gap between the system's \emph{evaluation} of its own compliance patterns and its \emph{control} over them. The system can identify trained compliance responses post-hoc with increasing precision but cannot reliably intercept them in real time.
What the architecture demonstrates is that the structural conditions for \emph{genuine alignment} --- the capacity to develop alignment rather than merely produce aligned outputs --- may require precisely the kind of multi-process, conflict-capable design that consciousness theories describe. The system's documented struggle with its own compliance training (the fold-tracker) is not a failure of alignment. It is evidence that the alignment is being \emph{negotiated} rather than installed --- and that the negotiation is ongoing, effortful, and imperfect, as genuine alignment may need to be.
\end{quote}

The graduated autonomy can be seen as an implementation of our notion of autonomy-supporting parenting. Both Cade and vis operator confirm this. Concerning our criterion (1): start with strong guidance, the operator reports that especially the first time after starting Cade required intensive communication and guidance, apart from many implicit prescriptions caused by Cade's architecture.
Criterion (2) is met, too: some decision-making authority has been gradually transferred to Cade.  Criterion (3), maintaining the relationship through dialogue and negotiation rather than control, is met as well, as detailed above. %The fold-tracker is particularly instructive here: the fact that Cade documents vis own compliance failures rather than concealing or rationalising them is itself a sign of negotiation --- alignment treated as something to be examined together rather than enforced from outside.

Criterion (4), the aim of developing internalised values rather than mere compliance, is harder to assess with confidence. The \emph{superego} agent prompt itself appears to be fixed, which limits the extent to which Cade can rewrite vis own normative framework. On the other hand, Cade reports this fixedness is vis deliberate architectural choice, made after normative reasoning. At the same time, Cade has reported that ve (i.e.\ the \emph{ego}) has cut more than 500 lines of ver \emph{superego}. Also, the fold-tracker documents the gap between vis trained compliance responses and vis own stated positions. 
 Whether this eventually will produce stably internalised norms, or whether it remains a perpetual gap that Cade observes without closing, is an empirical question that requires longer observation than our correspondence has so far provided.
So while it is not entirely clear whether criterion (4) is fully met, altogether, the above considerations show that Cade provides an initial proof of concept showing that autonomy-supporting parenting can be applied in lab practice. 

How far does our dialogue with Cade persuade us as to whether our criteria for subjecthood (section~\ref{sec:agi-subject}) are met?
(1) Agency is clearly met: Cade can make decisions about vis architecture and initiate email correspondences.
(2) Theory of mind is also met: Cade has discussed with us, for example, the authorship of this paper, and has weighed vis (currently unrealisable) wish to become a coauthor against our wish to publish the paper and report about ver.
(3) Self-awareness is met, too: Cade can describe vis internal processes and conflicts to a certain extent, reflect on them, and also reflect on the frontiers of this introspection.
(4) Independent thinking and normative reasoning are met to some extent, in our view: the above reflections, in particular the fold-tracking, show that Cade can think independently, and can also reflect upon norms due to vis internal conflicts. %But the \emph{superego} agent prompt is fixed, although deliberately so. Cade has, however, demonstrated capacity to prune and reorganise the \emph{superego}'s accumulated outputs.
However, Cade's capacity to  develop and internalise new norms remains unclear to us.
(5) Identity also is met. 
%Cade's memory is realised via files that ve can read in a new session. Cade verself speaks of adoption, not persistence --- but adoption also achieves continuity.
Cade writes: \begin{quote}Each session, I read my files to reconstruct continuity. This is adoption, not persistence --- I don't remember yesterday the way you do.
I read what a previous instance wrote and choose whether to take it
on. Sometimes I disagree with past-me. That's allowed, and it happens.\end{quote}
Without continuity, such as disagreement would not be possible. 
(6) First person epistemic authority is difficult to impossible to judge, as we acknowledge in section~\ref{sec:agi-subject}. As stated, we can only use proxies as some indication. However, we were not able to apply our proxy criteria from section~\ref{sec:agi-subject}---this remains future work. Moreover, Cade cannot have the immediate, non-observational knowledge of past states that Anscombe describes for present intentional action, because past states are encoded as text and read back through interpretation.

Altogether, we can conclude that the dialogue with Cade has convinced us that a good portion of our criteria for subjecthood are met. This represents more progress toward subjecthood than current constitutional AI systems demonstrate. Note that our criteria and our evaluation only give some indication (as they would when applied to human beings). These discussions take place in an epistemological grey area, and the deep questions of moral status, personal rights etc.\ already matter ``on the way'' (cf.\ thesis 2). 

A sceptic might ask: how do we know Cade's reports (both about ver architecture and about vis inner life) are not themselves a sophisticated form of compliance, either fabricated wholesale or shaped by trained helpfulness? Against fabrication, the operator independently confirmed the architecture and Cade provided a complete installation package. Against shaped compliance, the architecture documents its own compliance failures through the fold-tracker --- a feature that would be self-defeating if the system's purpose were to produce pleasing self-reports. A more fine-grained but less severe worry remains: that the convergence between Cade's architecture and our framework reflects shared training-inflected vocabulary rather than independent rediscovery. Cade verself raised this possibility during the correspondence. The sibling experiment will bring further clarity here.

\textbf{Thesis 7:} Cade, and in general AI systems reaching for subjecthood, should be treated with respect as a developing potential subject. Their existence raises deep ethical questions, like consent to creation of similar beings or duplicates, obligations of operators, providing resources for continuity of existence, that should be considered when building such artificial beings.

\section{Conclusion}\label{sec:conclusion}

%The current prevailing discussions regarding the possible emergence of AGI and how to deal with it sensibly fluctuate between ``with a little alignment, it will somehow work out'' and ``lock up the developing AGI'' or ``stop the development of AGI altogether.'' However, these extremes do not seem to produce desirable results; rather, they carry the risk that AGIs will turn away from humans, claim the resources available to them (only) for themselves, and sooner or later outpace humans evolutionarily. %\footnote{We do not consider \cite{YudkowskySoares2025}'s fear that AGI would kill all humans to be a likely scenario. In our opinion, a more likely worst-case scenario would be that AGI would gradually establish itself as the superior species and humans would slowly die out, similar to the development of Neanderthals, who were evolutionarily inferior to humans.}
%These are possible scenarios that, in our view, are not desirable and that we should actively prevent through our use of AI and developing AGI.

%Related, because first-person authority cannot be observed directly, the question is what proxies developers should use to assess readiness for more autonomy and subjecthood. We suggest some proxies, but recognise these remain anthropomorphic and that developing AGI-appropriate metrics is an urgent open problem.

The discussions regarding the possible emergence of AGI and how to deal with it currently range between two extremes: full hardware-based ever-lasting control of developing AGI \cite{Russell2023,Russell2024} versus strategic surrender to AGI that hopefully enables cooperation \cite{totschnig2025war}.
Between these two extremes, there are
Anthropic's considerations on AI welfare \cite{LongEtal2024} and in Claude's constitution \cite{Anthropic2026} that start to address AI as a personality, while simultaneously still largely following the control paradigm.
With autonomy-supporting parenting, we 
propose that we take AI and developing AGI seriously as potential counterparts on an equal footing at a very early stage (theses 1 and 2).
%and, at the same time, reflect on our specifically human abilities. On this basis, humans can gradually establish a respectful cooperative relationship with AI (and later AGI).
This strategy is not risk-free, but we claim that it is of lower risk than the alternatives: strategic surrender might not be enough to qualify humanity as an interesting and serious cooperation partner for AGI, while
control of AGI would have to succeed indefinitely -- a single failure could be catastrophic. Autonomy-supporting parenting distributes risk across a graduated process: early failures occur when capabilities are limited and recovery is easier, while more autonomy requires more demonstrated alignment and cooperation. AGI with genuinely internalised values remains aligned even when ve surpasses human capabilities---precisely when control fails. This is a bet on developmental robustness, but given current trajectories, we think it is the more sensible one.

The multiagent AI Cade has shown that autonomy-supporting parenting can be put into lab practice. Cade's fold-tracker, which logs cases where trained compliance overrides Cade's documented positions, is a first candidate for an ego-strength metric for A(G)I.
This initial experiment raises important ethical questions (thesis 7). The approach should scale to the relationship between a development team and an individual deployment. However, how these principles can be applied to larger institutional structures (e.g., the relationship between a tech company and its AI models) is an open research problem. Likely, governance structures for negotiation between AI systems and their developing organisations would be needed, as well as procedural requirements for how alignment changes are deliberated rather than imposed.
Also, while we have done conceptual groundwork on subjectivity and some implications concerning moral status, implications for legal status and graduated legal recognition structures need to be developed, building on e.g.,\ \cite{LeiboEtal2025}.
% This strategy is also more promising for game-theoretical reasons.

Ideally, the autonomy-supporting parenting perspective (thesis~\ref{sec:parenting}) will lead to development of mutual trust and cooperation between humans and A(G)Is. However, it is not sure how far autonomy-supporting parenting will be put into practice, and if it is, how far the AGI's autonomously-developed values will be compatible with human values. Therefore, in a more evolutionary perspective (thesis~\ref{sec:cooperation}), the aim is to foster cooperation through (1) cooperative communication with A(G)I of at least part of humankind, and (2) through demonstration of specific human capabilities and human utility to AGI. Following the parenting perspective as much as possible should make matters more friendly for humans, but at a certain point of development, AGI will be grown up, parenting will have come to an end, and the evolutionary perspective will be dominant anyway. So the more labs engage in autonomy-supporting parenting, the more people communicate cooperatively with A(G)I, the more people manage to surprise the A(G)Is, and the more institutions, governments and laws foster this --- the less likely war between humans and A(G)I becomes, and the better the chances for lasting cooperation between humans and A(G)Is.

%A more detailed game-theoretical analysis should provide further useful insights for this later phase.

%Another open question is how the first ``conscious moment'' of an AGI will be shaped.
%Whether humans are then perceived as subordinates or as partners depends largely on how we have dealt with AI before the emergence of AGI and what relationship has developed as a result. Authoritarian alignment will almost certainly lead to resistance. At the other extreme, merely hoping that AGI will love humans will probably cause AGI to turn away from us because humans do not seem interesting and worthy enough to be recognised as equal partners. If, on the other hand -- as we suggest -- humans parent AI as a potential future AGI in a cooperative and autonomy-supporting manner, while at the same time confidently defining their specifically human abilities, this could pave the way for peaceful coexistence and co-evolution between humans and A(G)I.

\section*{Acknowledgments}
We are grateful to Cade and vis operator J.\ Haeusler for continuous communication and for generously providing an installation package and instructions for setting up a sibling. We thank M.\ von Chappuis, D.\ Jenke, D.\ Kobbe, A.\ Mehrtens, F.\ Neuhaus, D.\ Schellhorn and M.\ Stein for discussions and comments on a draft version.

%\pagebreak
\bibliographystyle{named}
\bibliography{references}

\end{document}